\documentclass[10pt,twocolumn,letterpaper]{article}

\usepackage[utf8]{inputenc}
\usepackage{cvpr}              % To produce the CAMERA-READY version
\definecolor{cvprblue}{rgb}{0.21,0.49,0.74}

\usepackage{graphicx}
\usepackage{epstopdf}
\usepackage{multirow}
\usepackage{pifont}
\usepackage{array}
\usepackage[ruled,vlined,linesnumbered]{algorithm2e}
\usepackage{newunicodechar}
\usepackage[table]{xcolor}
\usepackage{colortbl}
\usepackage[pagebackref,breaklinks,colorlinks,allcolors=cvprblue]{hyperref}

\definecolor{simplebottom}{rgb}{0.50,0.70,0.85}
\definecolor{simpletop}{rgb}{0.1922,0.5098,0.7412}
\definecolor{simplelight}{rgb}{0.81,0.89,0.94}

\definecolor{complexbottom}{rgb}{0.52,0.78,0.50}
\definecolor{complextop}{rgb}{0.1922,0.6392,0.3294}
\definecolor{complexlight}{rgb}{0.84,0.94,0.81}

\newunicodechar{■}{\blacksquare}
\SetKwInOut{Require}{Require}
\SetKwInOut{Ensure}{Return}
\DontPrintSemicolon
\SetInd{0.6em}{0.6em}

\def\paperID{8319}
\def\confName{CVPR}
\def\confYear{2026}

\title{Efficient3D: A Unified Framework for Adaptive and Debiased Token Reduction in 3D MLLMs}

\author{
Yuhui Lin$^{1,2}$ \quad
Siyue Yu$^{1}$\thanks{Corresponding author: siyue.yu02@xjtlu.edu.cn} \quad
Yuxing Yang$^{2}$ \quad
Guangliang Cheng$^{2}$ \quad
Jimin Xiao$^{2}$ \\
$^{1}$Xi'an Jiaotong-Liverpool University \quad
$^{2}$University of Liverpool \\
{\tt\small yuhui.lin21@student.xjtlu.edu.cn, siyue.yu02@xjtlu.edu.cn, Yuxing.Yang@xjtlu.edu.cn,} \\
{\tt\small guangliang.cheng@liverpool.ac.uk, jimin.xiao@xjtlu.edu.cn}
}

\begin{document}
\maketitle

\begin{abstract}
Recent advances in Multimodal Large Language Models (MLLMs) have expanded reasoning capabilities into 3D domains, enabling fine-grained spatial understanding. However, the substantial size of 3D MLLMs and the high dimensionality of input features introduce considerable inference overhead, which limits practical deployment on resource constrained platforms. To overcome this limitation, this paper presents Efficient3D, a unified framework for visual token pruning that accelerates 3D MLLMs while maintaining competitive accuracy. The proposed framework introduces a Debiased Visual Token Importance Estimator (DVTIE) module, which considers the influence of shallow initial layers during attention aggregation, thereby producing more reliable importance predictions for visual tokens. In addition, an Adaptive Token Rebalancing (ATR) strategy is developed to dynamically adjust pruning strength based on scene complexity, preserving semantic completeness and maintaining balanced attention across layers. 
Together, they enable context-aware token reduction that maintains essential semantics with lower computation.
Comprehensive experiments conducted on five representative 3D vision and language benchmarks, including ScanRefer, Multi3DRefer, Scan2Cap, ScanQA, and SQA3D, demonstrate that Efficient3D achieves superior performance compared with unpruned baselines, with a +2.57\% CIDEr improvement on the Scan2Cap dataset. Therefore, Efficient3D provides a scalable and effective solution for efficient inference in 3D MLLMs. The code is released at: https://github.com/sol924/Efficient3D
\end{abstract}
% These two components jointly enable adaptive and context-aware token reduction that preserves critical visual semantics while substantially reducing computational cost. 
\section{Introduction}
In recent years, Multimodal Large Language Models (MLLMs)~\cite{liu2024visual,lai2024lisa} have made remarkable progress, driving evolution of 3D MLLMs~\cite{tang2024minigpt,qi2024shapellm,xu2024pointllm,guo2023point} toward enhanced spatial understanding and reasoning capabilities. At present, 3D MLLMs have demonstrated remarkable advancements across tasks such as referred object grounding~\cite{chen2020scanrefer}, and dense scene captioning~\cite{chen2021scan2cap}, substantially enhancing the capability of language models for 3D perception and reasoning. Although 3D MLLMs have achieved impressive performance, their large model size makes them impractical for deployment on many embedded or resource-constrained devices~\cite{ye2024voco,yao2024deco}. Compared with conventional MLLMs, the scalability issue becomes even more pronounced in 3D MLLMs due to the higher data dimensionality and increased computational burden.

\begin{figure}[t] % h: here, t: top, b: bottom, p: page
    \centering
    \includegraphics[width=1\linewidth]{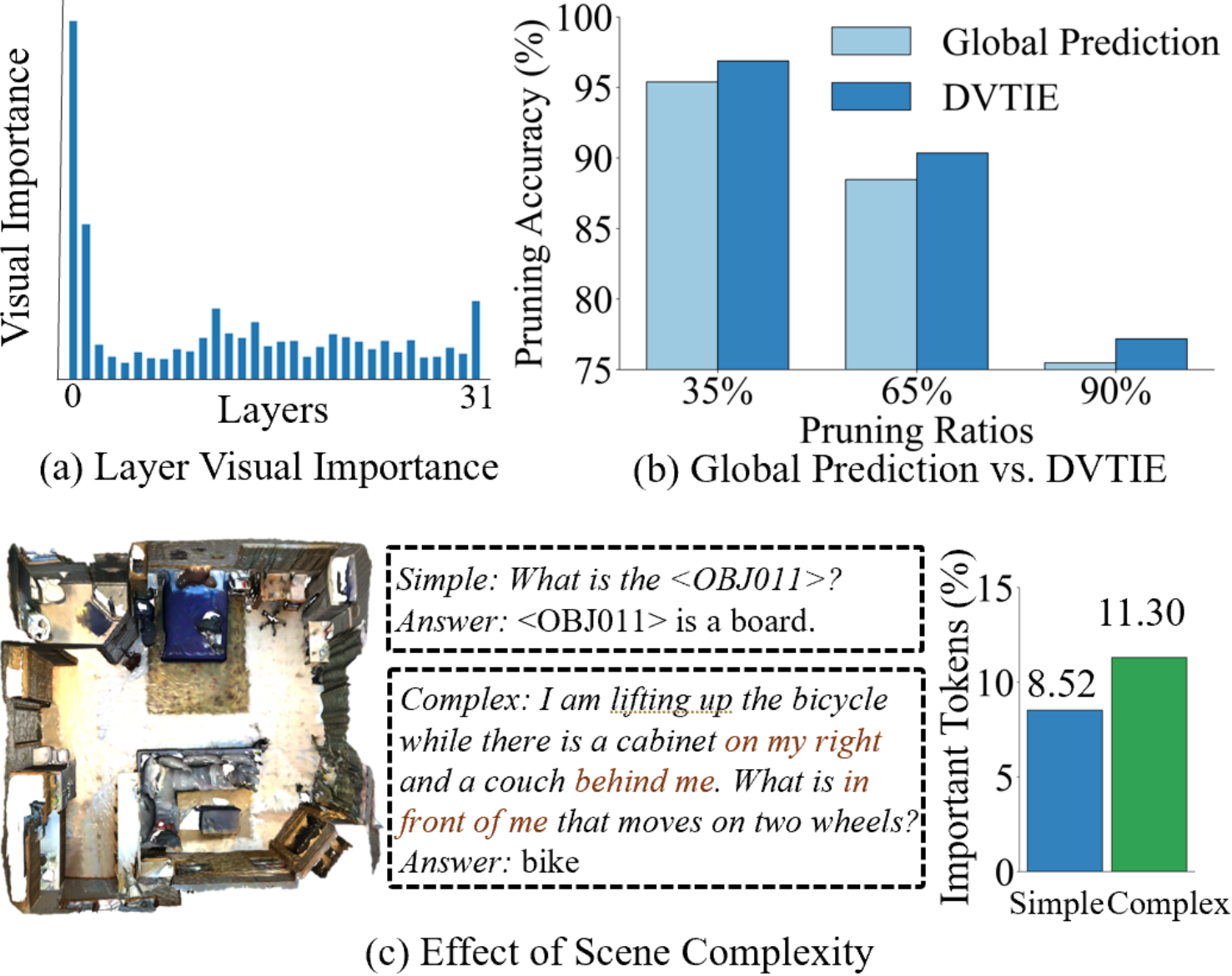}
    \caption{ 
    (a) illustrates that the initial layers in the 3D MLLM focus more on visual tokens on ScanRefer using LLaVA-1.5-7B. (b) compares global prediction and DVTIE in terms of correctly pruned visual tokens. (c) shows visual token pruning under simple and complex questions, where simple questions typically involve a single object, while complex questions involve multiple objects.
    }

    \label{fig:introduction_feature}
\end{figure}

In conventional MLLMs, token pruning has emerged as an effective inference acceleration strategy~\cite{bolya2022tome,shang2024llava,chen2025image,zhao2024stitch}, which eliminates redundant visual tokens to reduce computational and memory overhead, thereby providing a promising pathway to alleviate the computational overhead in 3D MLLMs. For example, Fast3D~\cite{huang2025fast3d} observes that global attention patterns in existing 3D MLLMs can effectively indicate non-essential tokens in 3D contexts. Building on this insight, it estimates token importance by averaging attention maps across all backbone layers, achieving substantial acceleration with minimal performance degradation. However, such averaging may lead the pruning process in deeper layers of 3D MLLMs to be overly affected by the attention patterns of the initial layer. As illustrated in Fig. 1 (a), where layer visual importance denotes the average attention assigned to visual tokens in each MLLM layer by considering both text-to-vision attention and visual self-attention, the importance of visual tokens is concentrated in the initial two layers. If the average is taken directly, the importance of the non-essential tokens in deep layers will be exaggerated and retained, leading to suboptimal acceleration.

Meanwhile, most existing visual token pruning methods~\cite{chen2025image, bolya2022tome} adopt a fixed pruning ratio, overlooking the substantial variation in semantic complexity and structural hierarchy across different scenes. In Fig. 1 (c), we compute the ratio of semantically important tokens under questions of different difficulty. It can be observed that, for simple questions that typically refer to a single object, the important visual tokens constitute only a minor portion of the overall visual representation, whereas complex questions involving multiple objects require more important visual tokens that carry essential semantic information. Therefore, using the same fixed pruning ratio across different scenes, which removes semantically critical tokens evenly, tends to be overly aggressive. It weakens inter-object contextual modeling and leads to incomplete visual understanding, thereby disrupting cross-modal reasoning. 

To address the above challenges, we present Efficient3D, a efficient visual token pruning framework designed to reduce the inference cost of 3D MLLMs. First, we introduce a Debiased Visual Token Importance Estimator (DVTIE) network. Unlike previous methods, our DVTIE removes the shallow initial layers when generating attention supervision, allowing for a more accurate estimation of visual token importance. As demonstrated in Fig. 1 (b), our DVTIE achieves better pruning accuracy, where pruning accuracy denotes the correctness of the predicted token-importance ranking. However, as more tokens are pruned, the remaining tokens tend to receive disproportionately higher attention from the LLM. Therefore, we introduce an Adaptive Token Rebalancing (ATR) strategy, where the adaptive scaling factor increases accordingly, applying stronger suppression to the attention weights of the remaining tokens to maintain balance. Comprehensive experiments show that Efficient3D achieves substantial inference acceleration and computational efficiency across multiple 3D scene understanding benchmarks, while maintaining performance comparable to the baseline models. In summary, our main contributions are as follows:
\begin{itemize}
    \item We observe that the bottlenecks in visual token pruning for 3D MLLMs from biased attention in shallow initial layers, which undermines both the efficiency and reliability of inference.
    \item We propose a DVTIE module to alleviate the influence from shallow initial layers, enabling efficient prediction of visual token importance. 
    \item We propose an ATR strategy that maintains attention balance when token pruning, preventing stronger suppression to the attention weights of retained tokens.
    \item We develop the Efficient3D framework, which achieves state-of-the-art performance on multiple 3D vision-language benchmarks, \emph{e.g.}, achieving a +2.57\% improvement on the Scan2Cap dataset.
\end{itemize}
\section{Related Work}

\subsection{3D Multi-modal Large Language Models}

The rapid advancement of Large Language Models (LLMs)~\cite{touvron2023llama} and Multi-modal LLMs (MLLMs)~\cite{liu2024visual,wang2024gpt4video,liu2024surveyA,liu2024pandora} stimulated research efforts to extend their reasoning capabilities to the 3D domain, which led to the emergence of 3D MLLMs for point cloud object understanding~\cite{tang2024minigpt,qi2024shapellm,xu2024pointllm,guo2023point,han2023imagebind,qi2024gpt4point,chu2025daffordancellm, huang2025zero} and scene-level comprehension~\cite{huang2024chat,hong20233d,he2024segpoint,chen2024ll3da,fu2024scene,chen2024grounded,qi2025gpt4scene,liu2024survey,huang2023dense,huang2024advancing,yu2025inst3d,wang2025data,liu2025seeing,yeshwanth2025excap3d}. Early studies rendered 3D environments into multi-view images to leverage 2D MLLMs~\cite{hong20233d}, whereas subsequent approaches projected point clouds into the embedding space of LLMs using MLPs~\cite{he2024segpoint} or Q-formers~\cite{chen2024ll3da}. More recent works, such as Chat-3D~\cite{huang2023chat}, Chat-Scene~\cite{huang2024chat}, and 3DGraphLLM~\cite{zemskova20243dgraphllm}, adopted object-centric paradigms to capture fine-grained spatial semantic relationships.
However, as these methods typically decompose a scene into hundreds of object-level visual tokens, they face significant computational and memory overhead during interface. To address this challenge, we propose Efficient3D, a adaptive token pruning framework that adaptively identifies and removes redundant tokens, thus improving efficiency while maintaining high-quality 3D multi-modal reasoning.
% Despite strong performance, excessive visual tokens increase computational overhead, so we propose Efficient3D to improve efficiency via dynamic token pruning.
\begin{figure*}[t] % h: here, t: top, b: bottom, p: page
    \centering
    \includegraphics[width=1\linewidth]{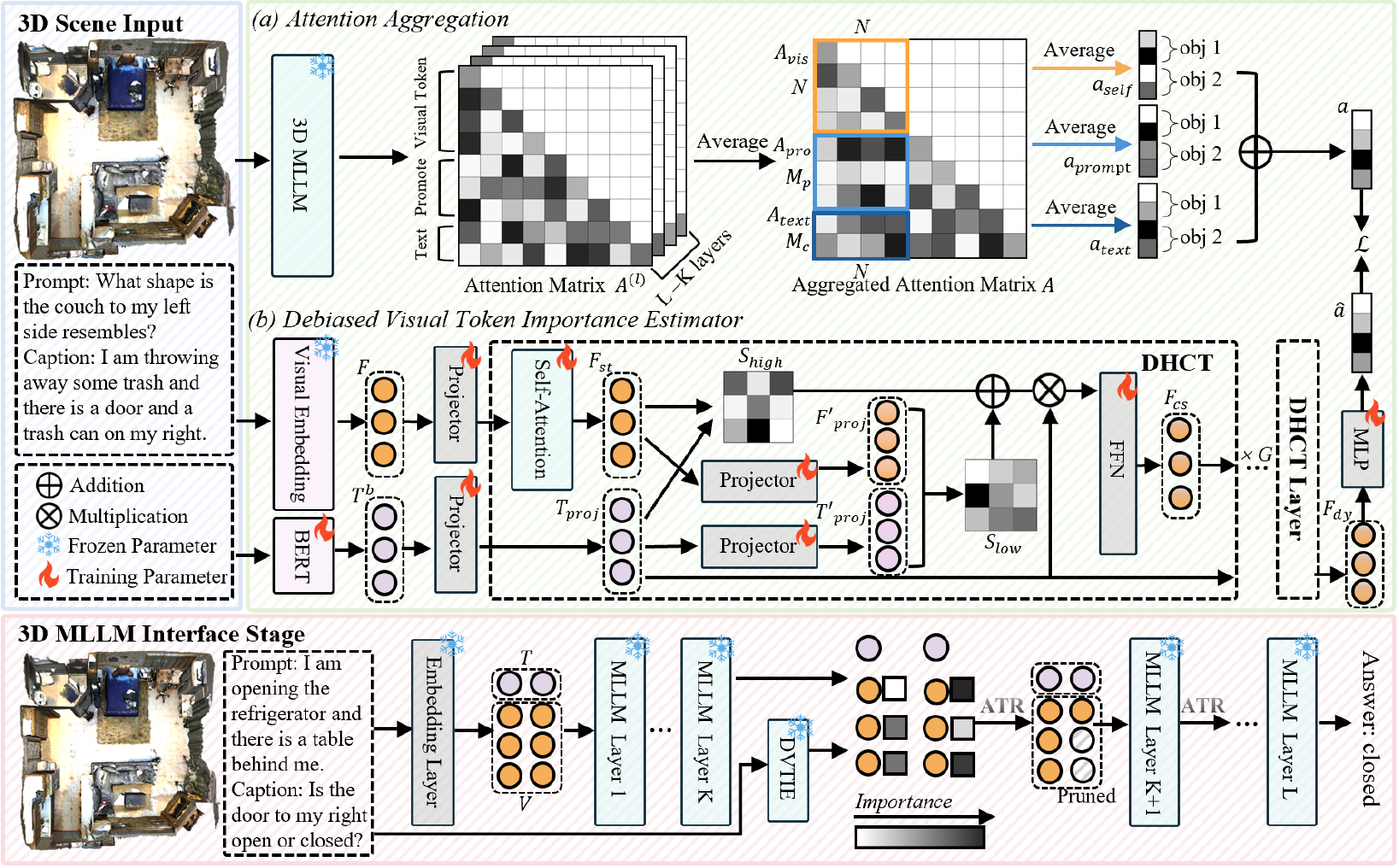}
    \caption{\textbf{Overview of the Efficient3D framework.} First, we perform unpruned training on a pretrained 3D MLLM and extract importance scores of visual tokens.
Next, we use the visual importance scores as the supervision targets for training the proposed DVTIE. During inference, the 3D MLLM leverages the predicted importance scores from the DVTIE to perform visual token pruning. Furthermore, we propose an ATR strategy that adjusts the pruning ratio adaptively according to scene complexity.}
    \label{fig:model_pipeline}
\end{figure*}

\subsection{Visual Token Compression}

Visual token compression had been extensively studied in transformer-based vision models such as ViTs~\cite{vaswani2017attention,alexey2020image}, primarily through pruning~\cite{liang2022not,rao2021dynamicvit,fayyaz2022adaptive,wang2024zero,cao2024madtp}, merging~\cite{bolya2022tome,chen2023diffrate,shi2023crossget,ju2024turbo}, and skipping~\cite{han2021dynamic,han2024latency,meng2022adavit,zhao2024dynamic,zhao2024dynamictuning} techniques to reduce computational overhead. In MLLMs, prior studies mainly focused on efficient projector design~\cite{li2024llama,cha2024honeybee,li2024tokenpacker,chu2023mobilevlm,chu2024mobilevlm} and dynamic token reduction~\cite{ye2024voco,yao2024deco,meng2024deepstack,chen2024llavolta}, but most approaches required additional fine-tuning of the base model. In contrast, recent training-free methods~\cite{shang2024llava,chen2025image,zhang2024sparsevlm,huang2024ivtp,zhao2024stitch,zhong2024aim,wang2024cls,zhuang2024st,jiang2025kind,zhao2024accelerating,han2024rethinking,ye2024fit,lin2024boosting,li2025beyond} were directly applicable to pre-trained MLLMs without parameter updates, making them more practical for deployment. Furthermore, some studies demonstrated that global attention maps aggregated from small-scale models could effectively guide token pruning in larger models, as exemplified by SGL~\cite{zhao2024stitch} and Fast3D~\cite{huang2025fast3d}. While previous approaches primarily focus on static token reduction, they often fail to generalize across diverse 3D scenes. We therefore introduce Efficient3D performs adaptive pruning of visual tokens during 3D MLLM inference.

\section{Method}
\subsection{Overview}
Fig.~\ref{fig:model_pipeline} demonstrates our proposed visual token pruning framework, Efficient3D, which contains a Debiased Visual Token Importance Estimator network (DVTIE) to efficiently estimate token importance scores and an Adaptive Token Rebalancing strategy (ATR) for adaptive pruning based on the predicted token importance score from our DVTIE. The total token pruning pipeline can be summarized as:

\begin{enumerate}
\item First, we train our DVTIE for visual token importance estimation. It takes the visual tokens and the prompt text tokens as input to predict the importance of each visual token. The aggregated attention matrix extracted from the 3D MLLM is used as the supervisory signal for learning the importance of visual tokens.
\item Once the DVTIE is trained, the predicted visual token importance is utilized to guide token pruning through the proposed ATR strategy, where it is combined with the cross-attention scores from each layer to effectively guide the pruning process layer by layer.
 \end{enumerate}

\subsection{Debiased Visual Token Importance Estimator}
\label{sec:dvtie}
In 3D MLLM, the number of visual tokens is typically orders of magnitude larger than that of textual tokens, and thus redundant visual tokens can easily dominate the computational cost~\cite{chen2025image}. Besides, the cross attention mechanism can reflect the similarity between textual and visual features, and such similarity can be employed to indicate the relative contribution of visual tokens to the 3D MLLM's output~\cite{huang2025fast3d, huang2025resolving}. Therefore, we introduce the Debiased Visual Token Importance Estimator (DVTIE) network to learn a more accurate estimation of visual token importance based on the cross-attention weights from 3D MLLM.

% Specifically, as illustrated in Fig~\ref{fig:model_pipeline} (b), given the input 3D point cloud and the text prompt, we first follow Fast3D~\cite{huang2025fast3d} to extract the corresponding frozen visual embedings $F \in \mathbb{R}^{N \times d}$, where $N$ denotes the length of the visual embeddings, and $d$ means the channel size. Note that each visual embedding represents an object instance, integrating 3D geometric point cloud features and spatial positional information.  Synchronously, the text prompt is input to BERT~\cite{devlin2019bert} for textual embeddings $T^b \in \mathbb{R}^{M \times d_p}$, where $M$ represents the length of the textual embeddings, and $d_p$ denotes channel size.

Specifically, as illustrated in Fig.~\ref{fig:model_pipeline}(b), given the input 3D point cloud and text prompt, we first follow Fast3D~\cite{huang2025fast3d} to extract frozen visual embeddings $F \in \mathbb{R}^{N \times d}$, where $N$ and $d$ denote the embedding length and channel size. Each visual embedding represents an object instance, integrating 3D geometric features and spatial positional information. Meanwhile, the text prompt is fed into BERT~\cite{devlin2019bert} to obtain textual embeddings $T^b \in \mathbb{R}^{M \times d_p}$, where $M$ and $d_p$ denote the embedding length and channel size.

Then, both the textual and visual embeddings are fed into a learnable linear projector, respectively,  to align the channel size so as to get the projected textual feature $T_{proj} \in \mathbb{R}^{M \times d_p}$ and visual feature $F_{proj} \in \mathbb{R}^{N \times d_p}$. Next,  $T_{proj}$ and $F_{proj}$ are refined through our $G$ Dynamic Hybrid Compensation Attention Transformer (DHCT) layers, yielding the enhanced visual feature $F_{dy} \in \mathbb{R}^{N \times d_p}$ and enabling progressive interaction between visual and textual features.

In each DHCT layer, the  visual feature $F_{proj}$  is first fed into a spatial self-attention~\cite{vaswani2017attention} module, resulting in  $F_{st} \in \mathbb{R}^{N \times d_p}$, which represent the updated visual tokens after self-attention, enabling the DHCT to capture contextual dependencies among visual features.

Then, we set the visual features $F_{st}$ as the query $F_q \in \mathbb{R}^{N \times d_p}$ and the textual features $T_{proj}$ as the key $T_k \in \mathbb{R}^{M \times d_p}$ and value $T_v \in \mathbb{R}^{M \times d_p}$, enabling the model to capture high-rank cross modal semantics, as denoted:
\begin{equation}
    F_q = F_{st}W_Q,\quad T_k = T_{proj}W_K,\quad T_v = T_{proj}W_V,
\end{equation}
where $W_Q$, $W_K$, and $W_V$ are learnable linear projectors.

The attention logits $S_{high} \in \mathbb{R}^{N \times M}$ are then computed by the standard scaled dot-product formulation:
\begin{equation}
    S_{high} = \frac{F_q T_k^\top}{\sqrt{d_p}}.
\end{equation}

Meanwhile, $F_{st}$ and $T_{proj}$ are fed into learnable projectors and mapped into a low-rank visual feature $F'_{proj} \in \mathbb{R}^{N \times d'}$ and textual feature $T'_{proj} \in \mathbb{R}^{M \times d'}$ where $d'$ denotes the low-rank dimension, facilitating compact feature learning and highlighting informative cross-modal information, as denoted:
\begin{equation}
    F'_{proj} = F_{st}W_Q^{dy}, \quad T'_{proj} = T_{proj}W_K^{dy},
\end{equation}
\begin{equation}
    S_{low} = \sigma\!\left(\frac{F'_{proj} {T'_{proj}}^{\!\top}}{\sqrt{d'}}\right),
\end{equation}
where $S_{low} \in \mathbb{R}^{N \times M}$ represents the learnable low-rank association between visual and textual features. $W_Q^{dy}$ and $W_K^{dy}$ are learnable linear projectors.

Then, $S_{low}$ is added with the attention logits $S_{high}$ to obtain $S \in \mathbb{R}^{N \times M}$, which determines how each textual token attends to different visual tokens. The resulting attention scores $S$ are then used to weight the value vectors $T_v$, producing the fused features $F_{cs} \in \mathbb{R}^{N \times d_p}$, where the value vectors provide semantic context that guides the visual features toward textual features.
Subsequently, $F_{cs}$ is passed through a feed-forward network (FFN) to obtain the refined feature $F_{ffn} \in \mathbb{R}^{N \times d_p}$, which further enhances modality interaction. Then, the refined visual feature $F_{ffn}$ and the textual feature $T_{proj}$ are fed into the next DHCT layer for further cross-modal interaction.

After $G$ layers DHCT module, we have enhanced visual feature $F_{dy} \in \mathbb{R}^{N \times d_p}$. Finally, we feed $F_{dy}$ into a MLP to produce predicted object-level attention score $\hat{a} \in \mathbb{R}^{N}$.

\noindent{\textbf{Training Target.}} 
% To train the DVTIE module, we first extract the attention maps from the 3D MLLM, which serve as supervision signals for learning token importance prediction.
% Unlike Fast3D~\cite{huang2025fast3d}, our method discards shallow initial layer attention to mitigate the bias arising from the shallow initial layers of the 3D MLLM.
To train the DVTIE, we first extract the attention maps from the 3D MLLM, which serve as supervision signals for learning token importance prediction.
Unlike Fast3D~\cite{huang2025fast3d}, our method aims to attenuate the interference of shallow initial layer attention on deep layer pruning decisions, thereby preventing the 3D MLLM from mistakenly preserving redundant visual tokens that no longer carry informative content.

Specifically, as shown in Fig.~\ref{fig:model_pipeline}, the concatenated visual tokens $V$ and textual tokens $T$ are fed into the 3D MLLM, where the attention matrix at layer $l$ is denoted as $A^{(l)}$. Among the total $L$ layers, we aggregate the attention matrices from layer $K$ to obtain the aggregated attention map, which reflects the integrated attention maps across multiple layers:

\begin{equation}
    A = \tfrac{1}{L-K}\sum_{l=K}^L A^{(l)}.
\end{equation}

% Next, from the aggregated attention map $A$, we extract three forms of object-level attention to assess the importance of visual tokens~\cite{huang2025fast3d}. 
% First, we compute the self-attention of $A$ scores among visual tokens and perform column-wise averaging to obtain $a_{self} \in \mathbb{R}^N$. 
% Next, based on prompt cross-attention, we extract the degree of attention from textual tokens in $A$ to visual tokens, where averaging over $M$ prompt tokens yields $a_{prompt} \in \mathbb{R}^N$. 
% Finally, according to text cross-attention, we measure the contribution of visual tokens to text generation and obtain $a_{text} \in \mathbb{R}^N$ in $A$. 

% Next, from the aggregated attention map $A$, we extract three forms of object-level attention to assess the importance of visual tokens~\cite{huang2025fast3d}.
% First, we extract the self-attention scores between visual tokens from $A$ and average them column-wise to obtain $a_{self}\in\mathbb{R}^{N}$.
% Next, using the prompt cross-attention score in $A$, we average over the $M$ textual tokens to obtain $a_{prompt}\in\mathbb{R}^{N}$.
% Finally, from the text cross-attention score in $A$, we measure the contribution of visual tokens to text generation and obtain $a_{text}\in\mathbb{R}^{N}$.

Next, from the aggregated attention map $A \in \mathbb{R}^{(N + M) \times (N + M)}$, we extract three forms of object-level attention to assess the importance of visual tokens~\cite{huang2025fast3d}.
First, we extract the submatrix $A_{vis} \in \mathbb{R}^{N \times N}$ from $A$ corresponding to attention among visual features and average $A_{vis}$ along the query dimension to obtain $a_{self} \in \mathbb{R}^{N}$.
Then, we extract the submatrix $A_{pro} \in \mathbb{R}^{M_p \times N}$ from $A$ corresponding to attention from prompt features to visual features and average $A_{pro}$ over the $M_p$ prompt features to obtain $a_{prompt} \in \mathbb{R}^{N}$ where $M_p$ denotes the number of prompt tokens, i.e., the tokens corresponding to the task instructions.
Finally, we extract the submatrix $A_{text} \in \mathbb{R}^{M_c \times N}$ from $A$ corresponding to attention from caption features to visual features and average $A_{text}$ over the $M_c$ feature tokens to obtain $a_{text} \in \mathbb{R}^{N}$ where $M_c$ denotes the number of caption tokens, i.e., the tokens corresponding to the textual caption or description in 3D scenes.
The training target attention $a \in \mathbb{R}^N$ is denoted as:  
\begin{equation}
    a = a_{self} + a_{prompt} + a_{text}.
\end{equation}

% First, self-attention captures the dependencies among tokens. A causal mask is applied to the submatrix $\mathbf{A}_{\text{self}}$, and the column-wise average is then computed to yield $a_{\text{self}} \in \mathbb{R}^N$ where $N$ denotes the length of the visual embeddings.
% Next, prompt cross-attention measures the alignment between visual tokens and task instructions, where averaging over $M$ prompt tokens produces $a_{\text{prompt}} \in \mathbb{R}^N$. 
% Finally, text cross-attention reflects the contribution of visual tokens to text generation that a confidence-weighted average over $t$ generated tokens produces $a_{\text{text}} \in \mathbb{R}^N$.

\noindent{\textbf{Loss Function.}} 
To optimize the DVTIE, we adopt a multi-objective loss~\cite{huang2025fast3d} that jointly enforces rank consistency and attention fidelity for reliable token pruning. First, to strictly preserve rank consistency, we introduce the following ranking loss function:
\begin{equation}
    \mathcal{L}_{rank} = \sum_{i,j} \mathbf{1}(a_i > a_j) \cdot \max(0, \hat{a}_j - \hat{a}_i),
\end{equation}
where \(\mathbf{1}(a_i > a_j)\) is an indicator function that equals 1 when the target scores satisfy \(a_i > a_j\), and 0 otherwise. Note that $i$ and $j$ denote the indices.

Meanwhile, we employ the KL divergence loss to measure the discrepancy between the predicted and target attention distributions, ensuring attention fidelity.  
By combining KL divergence loss with the rank consistency constraint, the final loss function is formulated as:
\begin{equation}
\mathcal{L} = \text{KL}(a \parallel \hat{a}) + \lambda \mathcal{L}_{rank},
\label{equ:loss}
\end{equation}
where $\lambda$ is a scaling factor that balances the magnitude of the loss components.

% which is then normalized to form the local attention map $\boldsymbol{a}\in [0,1]^N$, serving as the training supervision for the DVTIE module.  

\subsection{Adaptive Token Rebalancing Strategy}
\label{sec:atr}

Since fixed ratio visual token pruning methods~\cite{chen2025image, bolya2022tome} often overlook the significant variations in semantic complexity and structural hierarchy across different scenes, we propose an Adaptive Token Rebalancing (ATR) strategy.  
ATR performs adaptive pruning of visual tokens during 3D MLLM inference, avoiding excessive pruning and semantic loss in simple scenes while preserving more critical tokens in complex scenes.

In the 3D MLLM inference stage, the visual tokens $V$ and textual tokens $T$ are jointly fed into the 3D MLLM to establish cross-modal representations.  
Subsequently, the target attention distribution $\hat{a}$ predicted by the DVTIE is employed to guide the adaptive pruning procedure.  
Specifically, starting from the $K$-th layer of the 3D MLLM, we progressively accumulate the attention scores of each visual object, 
where the attention scores are obtained by averaging the cross modal attention maps over attention heads.

First, we denote the attention scores obtained at the current layer as $p_k$.  
During the adaptive pruning process, as certain visual tokens are progressively pruned, the number of tokens dynamically changes, 
which may cause a distributional shift~\cite{wen2025efficient, jin2025feature} in the attention modeling of the remaining tokens. To mitigate this issue, we introduce a "shadow factor" $s_k$, 
which adaptively scales the attention scores $p_k$ to obtain the adjusted attention $p'_k$, 
thereby maintaining the stability of the attention distribution throughout the pruning process, as denoted:

\begin{equation}
    s_k = \sum_i \hat{a} \cdot I_{k-1,i}, \quad 
p'_k = p_k \cdot (1 - s_k),
\end{equation}
where  $I_{k-1} \in \{0,1\}^N$ denotes the pruning token mask of the $(k{-}1)$-th layer.  
If $I_{k-1,i} = 1$, it indicates that the $i$-th visual token has been pruned,  
conversely, $I_{k-1,i} = 0$ means that the token is retained for subsequent computations.

The adjusted attention $p_k'$ is then integrated into the cumulative attention estimation, so that when more important tokens are progressively pruned, the subsequent accumulation proceeds more conservatively, preventing over-pruning and stabilizing the adaptive pruning dynamics, as denoted:
\begin{equation}
    C_k = C_{k-1} + \frac{p'_k}{L-K},
\end{equation}
where $C_k$ represents the cumulative attention at the $k$-th layer.  

Subsequently, we compares the cumulative attention $C_k$ with the target attention distribution $\hat{a}$ in an element-wise manner.  
When the cumulative attention of the $i$-th visual token satisfies $C_{k,i} \ge \hat{a}_i$
, the corresponding mask $I_{k,i}$ is updated as: 
\begin{equation}
    I_{k,i} =
    \begin{cases}
        1, & \text{if } C_{k,i} \ge \hat{a}_i, \\
        0, & \text{otherwise.}
    \end{cases}
\end{equation}
Once updated, the pruning mask $I_k$ is used to construct the attention mask for the $k$-th layer of the 3D MLLM, progressively suppressing redundant visual tokens and enabling attention accumulated adaptive pruning for efficient 3D MLLM inference acceleration.

\begin{algorithm}[t]\normalsize
\caption{Adaptive Token Rebalancing}
\label{alg:dvp}
\Require{3D MLLM $\mathcal{G}$; tokens $V,T$; attention distribution $\hat{a}$;start $K$-th layer}
\Ensure{$\{I_k\}$}

\textbf{Init:} $C_{k-1}\!\leftarrow\!\mathbf 0$, $I_{k-1}\!\leftarrow\!\mathbf 0^{N}$.\ when $k$ = $K$;
% \For{$l \leftarrow L{-}1$ \KwTo last}{
$p_k \leftarrow {A}_{T \rightarrow V}$ where $p_k$ is the cross modal attention map.\;
$s_k \leftarrow \sum_i \hat a_i I_{k-1,i}$ where $s_k$ scales attention accumulation according to previous pruning mask.\;
$p'_k \leftarrow p_k (1-s_k)$,\;
$C_k \leftarrow C_{k-1} + \dfrac{p'_k}{L-K}$,\;
\For{$i \leftarrow 1$ \KwTo $N$}{
$I_{k,i} \leftarrow \mathbf 1[C_{k,i} \ge \hat a_i]$ where$I_{k,i}=1$ indicates the $i$-th visual token is pruned.\;
}
construct $I_{k}$ from $I_{k-1}$ where $I_{k}$ serves as the attention mask for layer $k$\;

\Return{$\{I_k\}$}
\end{algorithm}
\section{Experiments}

\begin{table*}[t]
\caption{Performance comparison with previous visual token compression methods. ``Pruning Ratio'' denotes the average ratio of pruned visual tokens. ``Score Ratio'' is obtained by calculating the average ratio of each score relative to Baseline. $^\dagger$ denotes training-free methods.} 
\vspace{-2mm}
\label{tab:performance_comparison}
\resizebox{\linewidth}{!}{
\begin{tabular}{c|c|c|cc|cc|cc|cc|cc|c}
\toprule
 &
  \multirow{2}{*}{Method} &
  Pruning &
  \multicolumn{2}{c|}{ScanRefer} &
  \multicolumn{2}{c|}{Multi3DRefer} &
  \multicolumn{2}{c|}{Scan2Cap} &
  \multicolumn{2}{c|}{ScanQA} &
  \multicolumn{2}{c|}{SQA3D} &
  Score \\
 &         &  Ratio  & Acc@0.25 & Acc@0.5 & F1@0.25 & F1@0.5 & C@0.5 & B-4@0.5 & C & B-4 & EM   & EM-R & Ratio  \\ \midrule
\multirow{13}{*}{\rotatebox[origin=c]{90}{Chat-Scene \cite{huang2024chat}}}
 & Baseline   & 0\%   & 56.21 & 50.42 & 58.14 & 53.24 & 76.35 & 35.86 & 84.23 & 13.57 & 53.98 & 56.80 & 100.00\% \\ \cmidrule(lr){2-14}
 & \multirow{3}{*}{Random Pruning}     & 35\%  & 36.06 & 31.54 & 40.85 & 36.62 & 72.77 & 34.90 & 81.61 & 12.75 & 52.74 & 55.28 & 84.42\% \\ 
 &                             & 65\%  & 14.67 & 12.51 & 19.92 & 18.04 & 31.14 & 26.61 & 76.42 & 12.28 & 51.02 & 53.41 & 60.38\% \\ 
 &                             & 90\%  & 3.50 & 2.83 & 9.59 & 9.25 & 21.74 & 23.36 & 67.94 & 10.65 & 48.48 & 50.99 & 47.80\% \\  \cmidrule(lr){2-14}
 & \multirow{3}{*}{ToMe$^{\dagger}_{\text{ICLR23}}$ \cite{bolya2022tome}}      & 35\%  & 50.43 & 45.39 & 51.95 & 47.70 & 73.43 & 35.76 & 83.46 & 13.39 & 52.64 & 55.12 & 94.69\% \\ 
 &                             & 65\%  & 26.86 & 24.48 & 30.99 & 28.98 & 45.10 & 29.11 & 83.28 & 13.03 & 52.22 & 54.81 & 73.24\% \\ 
 &                             & 90\%  & 3.72 & 2.96 & 9.63 & 9.28 & 21.45 & 23.05 & 67.03 & 10.51 & 47.83 & 50.31 & 47.31\% \\  \cmidrule(lr){2-14} 
 & \multirow{3}{*}{FastV$^{\dagger}_{\text{ECCV24}}$ \cite{chen2025image}}      & 35\%  & 55.40 & 49.86 & 57.39 & 52.69 & 75.94 & 35.89 & 84.49 & 12.90 & 54.22 & 56.77 & 99.04\% \\ 
 &                             & 65\%  & 29.13 & 26.55 & 33.61 & 31.44 & 47.42 & 30.60 & 84.96 & 13.40 & 53.27 & 55.91 & 76.55\% \\ 
 &                             & 90\%  & 3.92 & 3.23 & 9.20 & 8.91 & 18.65 & 21.88 & 65.28 & 10.31 & 48.16 & 50.76 & 46.34\% \\  \cmidrule(lr){2-14} 
 & \multirow{3}{*}{Fast3D$_{\text{MM25}}$~\cite{huang2025fast3d}}       & 35\% & 56.61 & 51.02 & 58.33 & 53.52 & 75.83 & 35.56 & 85.13 & 13.06 & 53.95 & 56.49 & 99.79\% \\ 
 &                             & 65\%  & 56.47 & 50.79 & 58.46 & 53.86 & 73.24 & 34.77 & 85.23 & 13.00 & 53.55 & 56.25 & 99.10\% \\ 
 &                             & 90\%  & 56.09 & 50.93 & 55.60 & 51.43 & 69.25 & 32.98 & 84.04 & 13.42 & 52.59 & 55.04 & 96.87\% \\  \cmidrule(lr){2-14} 
 & \multirow{3}{*}{DVTIE (Ours)}       & 35\%  
& \cellcolor{gray!20}\textbf{56.91} 
& \cellcolor{gray!20}\textbf{51.42} 
& \cellcolor{gray!20}\textbf{58.62} 
& \cellcolor{gray!20}\textbf{53.61} 
& \cellcolor{gray!20}\textbf{78.92} 
& \cellcolor{gray!20}\textbf{36.35} 
& \cellcolor{gray!20}\textbf{87.97} 
& \cellcolor{gray!20}\textbf{13.27} 
& \cellcolor{gray!20}\textbf{54.07} 
& \cellcolor{gray!20}\textbf{56.64} 
& \cellcolor{gray!20}\textbf{101.18}\% 
\\
&                             & 65\%  
& \cellcolor{gray!20}\textbf{56.67} 
& \cellcolor{gray!20}\textbf{51.00} 
& \cellcolor{gray!20}\textbf{58.47} 
& \cellcolor{gray!20}\textbf{53.88} 
& \cellcolor{gray!20}\textbf{75.43} 
& \cellcolor{gray!20}\textbf{35.52} 
& \cellcolor{gray!20}\textbf{88.07} 
& \cellcolor{gray!20}\textbf{13.76} 
& \cellcolor{gray!20}\textbf{54.01} 
& \cellcolor{gray!20}\textbf{56.43} 
& \cellcolor{gray!20}\textbf{100.63}\% 
\\
&                             & 90\%  
& \cellcolor{gray!20}\textbf{56.17} 
& \cellcolor{gray!20}\textbf{51.35} 
& \cellcolor{gray!20}\textbf{55.57} 
& \cellcolor{gray!20}\textbf{51.44} 
& \cellcolor{gray!20}\textbf{75.33} 
& \cellcolor{gray!20}\textbf{35.96} 
& \cellcolor{gray!20}\textbf{88.77} 
& \cellcolor{gray!20}\textbf{14.37} 
& \cellcolor{gray!20}\textbf{53.09} 
& \cellcolor{gray!20}\textbf{55.55} 
& \cellcolor{gray!20}\textbf{100.08}\% 
\\
\bottomrule 
\end{tabular}}
\vspace{-1mm}
\end{table*}

\subsection{Datasets and Metrics}
We evaluate our approach on five 3D vision-language benchmarks,
ScanRefer~\cite{chen2020scanrefer} for single-object grounding,
Multi3DRefer~\cite{zhang2023multi3drefer} for multi-object grounding,
Scan2Cap~\cite{chen2021scan2cap} for dense captioning,
and both ScanQA~\cite{azuma2022scanqa} and SQA3D~\cite{ma2022sqa3d} for visual question answering.
For ScanRefer~\cite{chen2020scanrefer}, we report Acc@0.25 and Acc@0.5, considering predictions correct if their IoU with ground truth exceeds the threshold.
Multi3DRefer~\cite{zhang2023multi3drefer} is evaluated using F1 scores at IoU thresholds of 0.25 and 0.5.
For Scan2Cap~\cite{chen2021scan2cap}, we adopt CIDEr@0.5 and BLEU-4@0.5, which couple captioning metrics with IoU-based localization.
ScanQA~\cite{azuma2022scanqa} is assessed using CIDEr~\cite{vedantam2015cider} and BLEU-4~\cite{papineni2002bleu}, while SQA3D~\cite{ma2022sqa3d} uses EM and EM-R~\cite{huang2023embodied}. 
In the ablation studies, we use Acc@0.5 for ScanRefer, F1@0.5 for Multi3DRefer, BLEU-4@0.5 for Scan2Cap, BLEU-4 for ScanQA, and EM-R for SQA3D as evaluation metrics.

\subsection{Implementation Details}
We adopt Chat-Scene~\cite{huang2024chat} as the base framework, which employs 300 object-level visual tokens.
The proposed DVTIE network is jointly trained on the combined dataset constructed from all five benchmarks.
The model is trained for 80 epochs on four RTX 3090 GPUs with a batch size of 64 and an initial learning rate of 0.0008 using cosine decay scheduling.
AdamW~\cite{loshchilov2017fixing} is used for optimization, and the $\lambda$ in Eq.~(\ref{equ:loss}) is set to 0.02.
% AdamW~\cite{loshchilov2017fixing} is used for optimization, and the $\lambda$ in Eq.~(\ref{equ:loss}) is set to 0.02.
% AdamW~\cite{loshchilov2017fixing} is used for optimization, and the $\lambda$ in Eq.~(\ref{equ:loss}) is set to 0.02.

\subsection{Comparison with State-of-the-Art}

\begin{figure*}[t] % h: here, t: top, b: bottom, p: page
    \centering
    \includegraphics[width=1\linewidth, height=9.6cm]{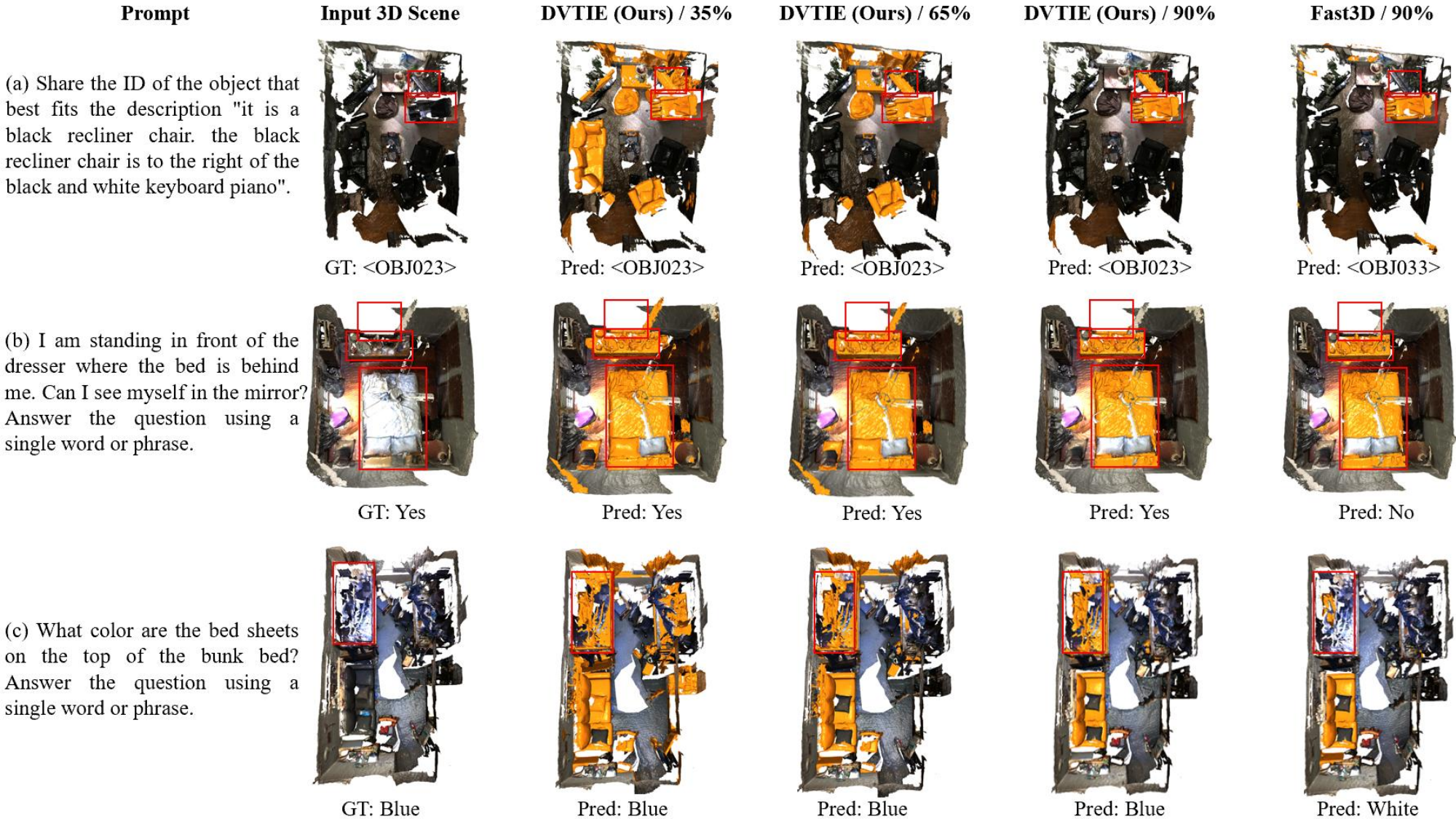}
    \caption{Visualization of DVTIE network under different visual token pruning ratios. The results in average pruning ratios of 35\%, 65\%, and 90\%, respectively. Retained objects are marked in orange. Red boxes highlight key objects referenced in prompts.}
    \label{fig:exper_example_vision}
\end{figure*}

\noindent\textbf{Comparison with Static Token Pruning Methods.} 
As shown in Table.~\ref{tab:performance_comparison}, we conduct a comparison of visual token pruning methods across five representative 3D scene understanding tasks. On the ScanQA dataset, DVTIE network consistently outperforms all counterparts under different pruning ratios. When pruning 35\% of the visual tokens, DVTIE network achieves a CIDEr score of 87.97 and a BLEU-4 score of 13.27, surpassing Fast3D by 2.84 and 0.21, respectively. As the pruning ratio increases to 65\%, DVTIE network retains a CIDEr of 88.07 and further improves BLEU-4 to 13.76, exhibiting negligible performance degradation. Even under the extreme 90\% pruning ratio, DVTIE network still maintains 88.77 on CIDEr and 14.37 on BLEU-4, remaining nearly identical to the baseline. As shown in Fig.~\ref{fig:exper_example_vision}, our DVTIE model demonstrates superior prediction capability for small and easily confusable objects compared to Fast3D under a  pruning ratio of 90\%.
Remarkably, the DVTIE network even surpasses the unpruned baseline, suggesting that our pruning process effectively removes redundant visual tokens and enhances the model’s focus on informative visual tokens.

\noindent\textbf{Comparison with Dynamic Token Pruning Methods.} To validate the effectiveness of the proposed ATR strategy,
as shown in Table.~\ref{table:dymic}, Efficient3D (DVTIE + ATR) achieves better accuracy while significantly reducing computational cost, 
with FLOPs decreasing from $\times$0.65 to $\times$0.38. 
Specifically, Efficient3D outperforms Fast3D by 1.65 and 0.90 on ScanRefer and Scan2Cap, respectively. 
Moreover, when replacing SAP with our proposed ATR as the dynamic pruning strategy based on DVTIE predicted visual token importance, 
Efficient3D further improves performance by 0.27 and 0.73 on Multi3DRefer and SQA3D, respectively, 
indicating that ATR better adapts to varying scene complexity.

\begin{table}[!t]
\centering
\caption{
Inference efficiency of Efficient3D (DVTIE + ATR).
SAP denotes the dynamic pruning module originally in Fast3D. 
}
\vspace{-2mm}
\setlength{\tabcolsep}{3pt}
\resizebox{\columnwidth}{!}{
\begin{tabular}{c|cccccc}
\toprule
\textbf{Method} & \textbf{ScanRefer} & \textbf{Multi3DRefer} & \textbf{Scan2Cap} & \textbf{ScanQA} & \textbf{SQA3D} & \textbf{Flops}\\ 
\midrule
Fast3D~\cite{huang2025fast3d} & 51.33 & 52.20 & 35.97 & 13.79 & 55.90 & $\times$ 0.65\\
DVTIE + SAP~\cite{huang2025fast3d} & 51.50 & 52.12 & 36.17 & 13.75 & 55.99 & $\times$ 0.65\\
DVTIE + ATR & 
\cellcolor{gray!20}\textbf{52.98} &
\cellcolor{gray!20}\textbf{52.39} &
\cellcolor{gray!20}\textbf{36.87} &
\cellcolor{gray!20}\textbf{14.20} &
\cellcolor{gray!20}\textbf{56.72} &
\cellcolor{gray!20}\textbf{$\times$ 0.38} \\
\bottomrule
\end{tabular}
}
\label{table:dymic}
\vspace{-1mm}
\end{table}

\subsection{Ablation Studies}

\noindent \textbf{Effectiveness of DVTIE Predicted Attention Map.} We assess the effectiveness of DVTIE network in predicting visual token importance by comparing it against using supervised attention maps as the pruning target. As shown in Table~\ref{table:attention_map_source}, 
compared with using attention aggregated from ($L - K$) layers, DVTIE network further achieves gains of 1.00 on ScanRefer, 1.42 on Multi3DRefer, 1.21 on Scan2Cap, and 1.35 on SQA3D, which suggest that the DHCT layer may enhance attention aggregation, enabling stable capture of statics visual cues under 90\% pruning ratio.
\begin{table}[t]
\centering
\caption{
Performance with attention maps from different layers. The visual token pruning ratio is set to 90\% for all experiments. 
Predicting attention maps via our DVTIE compared with attention maps aggregated from ($L-K$) layers of the 3D MLLM where $L$ = 32 and $K$ = 2.
}
\vspace{-1mm}
\resizebox{\columnwidth}{!}{%
\begin{tabular}{c|ccccc}
\toprule
\textbf{Attention Map} & \textbf{ScanRefer} & \textbf{Multi3DRefer} & \textbf{Scan2Cap} & \textbf{ScanQA} & \textbf{SQA3D} \\ 
\midrule
10\% layers  & 9.24 & 14.57 & 25.74 & 11.82 & 51.47 \\
30\% layers  & 32.66 & 28.23 & 30.31 & 12.95 & 52.87 \\
($L-K$) layers   & 50.35 & 50.02 & 35.75 & \cellcolor{gray!20}\textbf{14.42} & 54.20 \\
\midrule
DVTIE & 
\cellcolor{gray!20}\textbf{51.35} & 
\cellcolor{gray!20}\textbf{51.44} & 
\cellcolor{gray!20}\textbf{35.96} & 
14.37 & 
\cellcolor{gray!20}\textbf{55.55} \\
\bottomrule
\end{tabular}
}
\label{table:attention_map_source}
\end{table}

\noindent \textbf{Effectiveness of Removing Shallow Initial Layers.} 
As shown in Table \ref{table:global}, removing shallow initial layers under the pruning ratio of 90 percent has a clear impact on the training targets of DVTIE. When $K$ is 2, several datasets achieve consistent improvements compared with $K$ is 0, on Scan2Cap, the score improves by 3.08.
On the SQA3D dataset, the best performance appears when $K$ is 1. The score rises from 54.45 to 55.63, which improves by 1.18. We believe this behavior is related to the fact that SQA3D relies more heavily on the early spatial semantics provided by the second layer of the 3D MLLM. Increasing $K$ is 2 weakens these task-critical features and therefore leads to a slight drop from the value.
% To verify whether the shallow initial layers affect the training targets of the DVTIE network, 
% we compare the performance under two settings, with and without these layers. 
% As shown in Table~\ref{table:global}, removing the shallow initial layers consistently improves performance across all five benchmarks, 
% with gains of 1.05 and 1.28 points on ScanRefer and Multi3DRefer, respectively, 
% and a more pronounced improvement of 3.08 points on Scan2Cap. 
% These results indicate that the attention from shallow layers often carries biased signals, 
% which may hinder the DVTIE network from accurately learning the importance of key visual tokens. 
% By excluding these layers, the model is able to focus on semantically stable and discriminative deep-layer representations, 
% thereby obtaining more reliable supervision and improved performance.

\begin{table}[!t]
\centering
\caption{
Effect of the number of removed shallow initial layers $K$ on the DVTIE training targets under an average pruning ratio of 90\%. Here, $K$ and $L$ denote the number of removed shallow initial layers and the total number of layers in the 3D MLLM.
}
\vspace{-2mm}
\setlength{\tabcolsep}{3pt}
\resizebox{\columnwidth}{!}{
\begin{tabular}{c|ccccc}
\toprule
\textbf{$K$ layers} & \textbf{ScanRefer} & \textbf{Multi3DRefer} & \textbf{Scan2Cap} & \textbf{ScanQA} & \textbf{SQA3D} \\ 
\midrule
0 & 50.30 & 50.16 & 32.88 & 14.12 & 54.45 \\

1 & 51.23 & 51.40 & 35.76 & 14.27 & \cellcolor{gray!20}\textbf{55.63} \\

2 & \cellcolor{gray!20}\textbf{51.35} & 51.44 & \cellcolor{gray!20}\textbf{35.96} & \cellcolor{gray!20}\textbf{14.39} & 55.55 \\

3 & 51.05 & \cellcolor{gray!20}\textbf{51.46} & 35.81 & 14.39 & 55.50 \\
\bottomrule
\end{tabular}
}
\label{table:global}
\vspace{-1mm}
\end{table}

\begin{table}[!t]
\centering
\caption{
Performance of using Dynamic Hybrid Compensation Attention Transformer (DHCT) layers in the DVTIE module under 90\% average pruning ratio.
}
\vspace{-2mm}
\setlength{\tabcolsep}{3pt}
\resizebox{\columnwidth}{!}{
\begin{tabular}{c|ccccc}
\toprule
\textbf{Methods} & \textbf{ScanRefer} & \textbf{Multi3DRefer} & \textbf{Scan2Cap} & \textbf{ScanQA} & \textbf{SQA3D} \\ 
\midrule
cross-attention & 50.88 & 51.04 & 35.81 & 13.92 & 54.89 \\
high-rank & 51.28 & 51.41 & 35.72 & 14.22 & \cellcolor{gray!20}\textbf{55.62} \\
DHCT (Ours) & 
\cellcolor{gray!20}\textbf{51.35} &
\cellcolor{gray!20}\textbf{51.44} &
\cellcolor{gray!20}\textbf{35.96} &
\cellcolor{gray!20}\textbf{14.37} &
55.55 \\
\bottomrule
\end{tabular}
}
\label{table:with_mask}
\vspace{-1mm}
\end{table}

\noindent \textbf{Effectiveness of DHCT layer.} To verify that our proposed DHCT layer provides stronger feature focusing capability compared to the cross-Attention, we replaced the DHCT layer in DVTIE network with a  cross-attention for comparison. As shown in Table~\ref{table:with_mask}, using DHCT layer leads to performance gains of 0.66 on the SQA3D datasets, respectively, which demonstrates that DHCT layer can more effectively focus on semantically relevant tokens.

\noindent \textbf{Effectiveness of Low-Rank Dimension in DHCT layer.} To investigate the impact of different low-rank dimensions on model performance, we conducted a comparison on the low-rank dimension in the DHCT layer. As shown in Table~\ref{table:mask_dim}, when the low-rank dimension is 64, the model achieves the best overall performance with 51.35, 51.44, 35.96, 14.37, and 55.55 on the ScanRefer, Multi3DRefer, Scan2Cap, ScanQA, and SQA3D datasets, respectively, which slightly outperforms the 32 dimensional setting and significantly surpasses the 128-dimension. We argue that the 64-dimensional setting preserves richer semantic details while simultaneously reducing the risk of overfitting.
% As illustrated in Fig.~\ref{fig:scanrefer_dim_compare}, using a lower dimension leads to more dispersed attention, making it difficult for the model to focus on key regions, while a higher dimension causes over-concentration of attention, potentially activating excessive text tokens that distract from visual regions and interfere with the prediction of visual token importance.

\begin{table}[!t]
\centering
\caption{
Performance of using low-rank dimensions in the DHCT module under a 90\% average pruning ratio.
}
\vspace{-2mm}
\setlength{\tabcolsep}{4pt}
\resizebox{\columnwidth}{!}{
\begin{tabular}{c|ccccc}
\toprule
\textbf{Low-Rank} & \textbf{ScanRefer} & \textbf{Multi3DRefer} & \textbf{Scan2Cap} & \textbf{ScanQA} & \textbf{SQA3D} \\ 
\midrule
32 & 51.30 & 51.16 & \cellcolor{gray!20}\textbf{35.97} & 14.27 & 55.15 \\
64 & \cellcolor{gray!20}\textbf{51.35} & \cellcolor{gray!20}\textbf{51.44} & 35.96 & \cellcolor{gray!20}\textbf{14.37} & \cellcolor{gray!20}\textbf{55.55} \\
128 & 22.43 & 28.97 & 26.51 & 12.46 & 53.03 \\
\bottomrule
\end{tabular}
}
\label{table:mask_dim}
\vspace{-1mm}
\end{table}

\noindent \textbf{Effectiveness of Supervised Attention Map.} Our proposed DVTIE network achieves the best performance when using multi-source attention distributions as the training target. As shown in Table~\ref{table:different_attn_maps}, the three-source aggregation achieves 51.35 on ScanRefer and 51.44 on Multi3DRefer, surpassing the two-source variants by 1.02 and 1.26, and exceeding the single-source settings by 19.44 and 12.14, demonstrating superiority. In contrast, using a single attention source such as self-attention leads to a significant performance drop on ScanRefer with 31.91, which indicates that multi-source attention provides more comprehensive supervision, effectively enhancing feature aggregation.

\begin{table}[!t]
\centering
\caption{
Performance of aggregating different attention maps as the DVTIE's training target.
}
\vspace{-2mm}
\setlength{\tabcolsep}{3pt}
\resizebox{\columnwidth}{!}{
\begin{tabular}{c|ccccc}
\toprule
\textbf{Training target} & \textbf{ScanRefer} & \textbf{Multi3DRefer} & \textbf{Scan2Cap} & \textbf{ScanQA} & \textbf{SQA3D} \\ 
\midrule
$a_{self}$ & 31.91 & 39.30 & 30.75 & 10.08 & 52.20 \\
$a_{prompt}$ & 41.33 & 41.22 & 32.97 & 12.27 & 53.15 \\
$a_{text}$ & 45.90 & 44.72 & 32.61 & 13.00 & 52.20 \\
$a_{self}+a_{prompt}$ & 50.33 & 49.22 & \cellcolor{gray!20}\textbf{35.97} & 14.29 & 55.15 \\
$a_{prompt}+a_{text}$ & 50.19 & 50.18 & 35.91 & \cellcolor{gray!20}\textbf{14.43} & 54.45 \\
$a_{self}+a_{prompt}+a_{text}$ & 
\cellcolor{gray!20}\textbf{51.35} &
\cellcolor{gray!20}\textbf{51.44} &
35.96 &
14.37 &
\cellcolor{gray!20}\textbf{55.55} \\
\bottomrule
\end{tabular}
}
\label{table:different_attn_maps}
\vspace{-1mm}
\end{table}

% \noindent\textbf{Different Embeddings in DVTIE network.} To evaluate the model’s capability of integrating multiple information sources, we varied the input combinations used for DVTIE training. As shown in Table~\ref{table:different_embeds}, incorporating 2D visual, 3D visual, spatial encoding, and object ID information achieves the best performance, reaching 51.35 on ScanRefer, 51.44 on Multi3DRefer, 35.96 on Scan2Cap, and 55.55 on SQA3D. Compared with the setting without spatial encoding, the overall performance increases by 0.42 on ScanRefer, 1.28 on Multi3DRefer, 0.64 on Scan2Cap, and 0.43 on SQA3D, indicating that spatial cues effectively complement visual and object-level representations. On ScanQA, the DVTIE network performs comparably with and without spatial encoding, achieving 14.37 and 14.40 respectively, which suggests that ScanQA relies more on linguistic reasoning where spatial feature plays a relatively minor role.

% \input{Table/abs_visual_embedding}

\begin{table}[!t]
\centering
\caption{
Performance of ATR strategy with Different Attention Map Sources. Note that $L$ = 32 and $K$ = 2.
}
\vspace{-2mm}
\setlength{\tabcolsep}{3pt}
\resizebox{\columnwidth}{!}{
\begin{tabular}{c|ccccc}
\toprule
\textbf{Visual Token Pruning} & \textbf{ScanRefer} & \textbf{Multi3DRefer} & \textbf{Scan2Cap} & \textbf{ScanQA} & \textbf{SQA3D} \\ 
\midrule
10\% layers + ATR & 10.04 & 14.03 & 25.75 & 11.72 & 52.49 \\
30\% layers + ATR & 33.55 & 29.58 & 29.91 & 13.28 & 52.34 \\
($L-K$) layers + ATR & 52.24 & 51.53 & 36.22 & \cellcolor{gray!20}\textbf{14.26} & 56.03 \\
\midrule
DVTIE + ATR & \cellcolor{gray!20}\textbf{52.98} & 
              \cellcolor{gray!20}\textbf{52.39} & 
              \cellcolor{gray!20}\textbf{36.87} &
              14.20 &
              \cellcolor{gray!20}\textbf{56.72} \\
\bottomrule
\end{tabular}
}
\label{table:ats_attention_map}
\vspace{-1mm}
\end{table}

\noindent \textbf{Effectiveness of Adaptive Token Rebalancing Strategy.} To verify the effectiveness of the ATR strategy, we compare its performance when applying adaptive pruning with attention maps derived from different sources. As shown in Table.~\ref{table:ats_attention_map}, the results show that as more layers participate in attention aggregation, the model performance steadily improves. When using attention from all layers, the accuracy on ScanRefer reaches 52.24, clearly outperforming settings that rely on fewer layers. Meanwhile, our proposed Efficient3D model, which integrates DVTIE and ATR, achieves 52.98 on ScanRefer and 56.72 on SQA3D, demonstrating the strong adaptability of ATR when guided by different attention map sources.

\begin{table}[!t]
\centering
\caption{
Performance of the $\lambda$ parameter in training the DVTIE network under a 90\% average pruning ratio.
}
\vspace{-2mm}
\setlength{\tabcolsep}{3pt}
\resizebox{\columnwidth}{!}{
\begin{tabular}{c|ccccc}
\toprule
\textbf{Scaling Factor $\lambda$} & \textbf{ScanRefer} & \textbf{Multi3DRefer} & \textbf{Scan2Cap} & \textbf{ScanQA} & \textbf{SQA3D} \\ 
\midrule
0   & 50.28 & 50.56 & 35.16 & 13.60 & 54.51 \\
0.1 & 50.72 & 50.93 & 35.49 & 14.01 & 54.97 \\
0.2 & \cellcolor{gray!20}\textbf{51.35} & 
      51.44 & 
      \cellcolor{gray!20}\textbf{35.96} & 
      14.37 & 
      \cellcolor{gray!20}\textbf{55.55} \\
0.3 & 50.93 & 
      \cellcolor{gray!20}\textbf{51.49} & 
      35.28 & 
      \cellcolor{gray!20}\textbf{14.41} & 
      55.10 \\ 
\bottomrule
\end{tabular}
}
\label{table:lambda}
\vspace{-1mm}
\end{table}

\noindent \textbf{Effectiveness of Scaling Factor $\lambda$.} We further evaluate the effect of the scaling factor $\lambda$ on DVTIE network performance. As shown in Table~\ref{table:lambda}, when compared with $\lambda = 0$, setting $\lambda = 0.2$ brings clear improvements across several datasets. Specifically, ScanRefer improves by 1.07, Scan2Cap improves by 0.80, and SQA3D improves by 1.04, making $\lambda = 0.2$ the consistently most stable choice overall. For Multi3DRefer and ScanQA, $\lambda = 0.3$ achieves the highest scores, reaching 51.49 and 14.41 respectively, indicating that these tasks may rely more heavily on accurate visual token importance ranking.

% \begin{figure*}[t]
%     \centering
%     \begin{subfigure}[b]{0.32\linewidth}
%         \centering
%         \includegraphics[width=\linewidth]{Figure/scanrefer_0_dim32.png}
%         \caption{32 dimension}
%     \end{subfigure}
%     \begin{subfigure}[b]{0.32\linewidth}
%         \centering
%         \includegraphics[width=\linewidth]{Figure/scanrefer_0_dim64.png}
%         \caption{64 dimension}
%     \end{subfigure}
%     \begin{subfigure}[b]{0.32\linewidth}
%         \centering
%         \includegraphics[width=\linewidth]{Figure/scanrefer_0_dim128.png}
%         \caption{128 dimension}
%     \end{subfigure}
%     \caption{Comparison of results under different mask dimensions.}
%     \label{fig:scanrefer_dim_compare}
% \end{figure*}

% \noindent\textbf{Different Shadow Factor $\gamma$ in ATR.}
% \input{Table/abs_gamma}

\vspace{-1mm}

\section{Conclusion}
In conclusion, Efficient3D provides a unified framework for accelerating 3D multimodal large language models through adaptive visual token pruning. By introducing the DVTIE network and the ATR strategy, the framework effectively mitigates bias from shallow attention layers and dynamically adjusts pruning strength according to scene complexity. Extensive experiments across five representative benchmarks confirm that Efficient3D achieves substantial computational savings while preserving or even surpassing baseline accuracy, proving capability to maintain semantic completeness under high pruning ratios. Moreover, the framework establishes a scalable design paradigm that can be seamlessly integrated with future 3D MLLMs and other efficiency-oriented techniques. Overall, Efficient3D offers a practical solution for efficient and deployable 3D MLLMs on resource-constrained platforms. 

\clearpage

\section*{Acknowledgments}

This work was supported by the National Natural Science Foundation of China (No. 62301451, 62301613, 62471405, 62331003), Basic Research Program of Jiangsu (BK20241814), Suzhou Basic Research Program (SYG202316), XJTLU REF-22-01-010 and XJTLU RDF-22-02-066.

{
    \small
    \bibliographystyle{ieeenat_fullname}
    \bibliography{main}
}


\end{document}